\documentclass[runningheads]{llncs}

\usepackage{eccv}

\usepackage{eccvabbrv}

\usepackage{graphicx}
\usepackage{booktabs}
\usepackage{multirow}
\usepackage{pifont}

\usepackage[accsupp]{axessibility}  % Improves PDF readability for those with disabilities.

\usepackage[breaklinks,colorlinks,citecolor=eccvblue]{hyperref}
\usepackage{orcidlink}

\begin{document}

% ---------------------------------------------------------------
% TODO REVIEW: Replace with your title
\title{EAFormer: Scene Text Segmentation with Edge-Aware Transformers} 

% TODO REVIEW: If the paper title is too long for the running head, you can set
% an abbreviated paper title here. If not, comment out.
\titlerunning{EAFormer: Scene Text Segmentation with Edge-Aware Transformers}

% TODO FINAL: Replace with your author list. 
% Include the authors' OCRID for the camera-ready version, if at all possible.
\author{Haiyang Yu\orcidlink{0000-0003-2747-7338} 
\and Teng Fu\orcidlink{0009-0009-8413-2185}
\and Bin Li\thanks{Corresponding Author}\orcidlink{0000-0002-9633-0033}
\and Xiangyang Xue\orcidlink{0000-0002-4897-9209}
}

% TODO FINAL: Replace with an abbreviated list of authors.
\authorrunning{Haiyang Yu et al.}
% First names are abbreviated in the running head.
% If there are more than two authors, 'et al.' is used.

% TODO FINAL: Replace with your institution list.
\institute{
Shanghai Key Laboratory of Intelligent Information Processing\\ School of Computer Science, Fudan University\\
\email{\{hyyu20, libin, xyxue\}@fudan.edu.cn},
\email{tfu23@m.fudan.edu.cn}}

\maketitle

\begin{abstract}
  Scene text segmentation aims at cropping texts from scene images, which is usually used to help generative models edit or remove texts. The existing text segmentation methods tend to involve various text-related supervisions for better performance. However, most of them ignore the importance of text edges, which are significant for downstream applications. In this paper, we propose Edge-Aware Transformers, termed EAFormer, to segment texts more accurately, especially at the edge of texts. Specifically, we first design a text edge extractor to detect edges and filter out edges of non-text areas. Then, we propose an edge-guided encoder to make the model focus more on text edges. Finally, an MLP-based decoder is employed to predict text masks. We have conducted extensive experiments on commonly-used benchmarks to verify the effectiveness of EAFormer. The experimental results demonstrate that the proposed method can perform better than previous methods, especially on the segmentation of text edges. Considering that the annotations of several benchmarks (\textit{e.g.}, COCO\_TS and MLT\_S) are not accurate enough to fairly evaluate our methods, we have relabeled these datasets. Through experiments, we observe that our method can achieve a higher performance improvement when more accurate annotations are used for training. The code and datasets are available at \href{https://hyangyu.github.io/EAFormer/}{\textcolor{blue}{https://hyangyu.github.io/EAFormer/}}.
\end{abstract}

\section{Introduction}
\label{sec:intro}

In the last decade, scene text segmentation~\cite{dai2018fused,zhou2013scene,ren2022looking,tang2017scene} has gained significant traction, primarily due to advancements in deep learning. The objective of a text segmentation model is to accurately distinguish between foregrounds (text areas) and backgrounds (non-text areas) at the pixel level. Scene text segmentation plays a crucial role in various applications, such as document analysis~\cite{fujisawa1992segmentation,pack2023perceptual}, scene text image super-resolution~\cite{ma2022textsrnet,shu2023text}, scene understanding~\cite{ess2009segmentation} and text erasing~\cite{zdenek2020erasing,lyu2023fetnet,du2023modeling,conrad2021two}. For example, TEAN~\cite{shu2023text} introduces the text segmentation results as auxiliary information to better super-resolve scene text images.

In order to promote the development of scene text segmentation, various methods~\cite{wang2023textformer,andreini2021two,zu2023weakly} and datasets~\cite{xu2022bts,xu2021rethinking,bonechi2019coco_ts} have been proposed in recent years. Previous scene text segmentation methods tend to introduce text-related supervision, such as text or character recognition supervision, to improve performance. TexRNet~\cite{xu2021rethinking} proposes a pre-trained character discriminator to introduce the supervision of character recognition, which requires additional annotations of character-level bounding boxes. Similarly, PGTSNet~\cite{xu2022bts} designs a text perceptual discriminator to enhance the readability of segmentation results. In addition, both of them utilize various losses for better segmentation performance, which may make it challenging to select appropriate hyper-parameters to balance multiple losses. Recently, TextFormer~\cite{wang2023textformer} employs a recognition head to make the model focus on text details and improve its perception of texts. For the scene text segmentation task, there are several widely-used benchmarks, such as ICDAR13 FST~\cite{karatzas2013icdar}, COCO\_TS~\cite{bonechi2019coco_ts}, MLT\_S~\cite{bonechi2020weak}, Total-Text~\cite{ch2017total}, TextSeg~\cite{xu2021rethinking} and BTS~\cite{xu2022bts}. Although the samples of these datasets seem to be adequate for deep-learning-based models, the annotation quality of some datasets (\textit{e.g.}, MLT\_S) may not meet the desired standards, especially in text-edge areas. The annotations of these datasets are obtained through bounding-box supervision, which  cannot provide the same level of accuracy and precision as the datasets annotated by humans, such as TextSeg and BTS.

\begin{figure*}[t]
  \centering
  \includegraphics[width=1.0\linewidth]{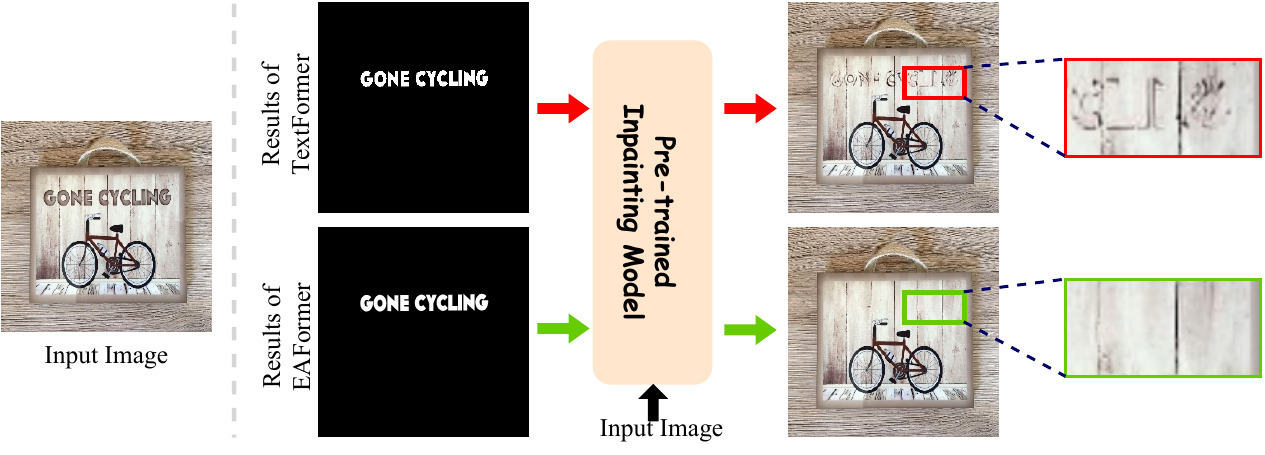}

   \caption{Results comparison of the downstream application (text erasing) with different text masks as input. More accurate segmentation at text edges is beneficial to the text erasing task since less text pixels are wrongly predicted and more background information is reserved for the inpainting model.}
   \label{fig:intro}
\end{figure*}

Although previous methods have achieved a certain performance improvement in text segmentation, they ignore the significance of text edges in practical applications. For instance, accurate text masks, especially in text-edge areas, can provide more background information to inpaint text areas in the text erasing task, as shown in Figure~\ref{fig:intro}. In experiments, we observe that traditional edge detection algorithms, such as Canny~\cite{canny1986computational}, can well distinguish text edges. To fully exploit the merits of traditional edge detection methods to improve the segmentation performance at text edges, in this paper, we propose Edge-Aware Transformers (EAFormer) for scene text segmentation. Specifically, EAFormer consists of three main modules: text edge extractor, edge-guided encoder and text segmentation decoder. The text edge extractor takes the scene images as input and predicts the text areas and edges. In this module, we adopt a light-weighted backbone to detect text areas and use the traditional edge detection algorithm Canny to obtain the edges of whole images. To alleviate the interference of edges in non-text areas, the masks of predicted text areas are used to filter out the edges of non-text areas. For the edge-guided encoder, we adopt the framework of SegFormer~\cite{xie2021segformer}, which is composed of four stages. At the first stage of this encoder, we additionally design a symmetric cross-attention sub-module, which aims at utilizing the filtered text edges to guide the encoder to focus more on text edges at the first stage. Finally, the outputs of the edge-guided encoder are fused and input into the text segmentation decoder to predict text masks.

To validate the effectiveness of EAFormer, we have conducted extensive experiments on six text segmentation benchmarks. The results demonstrate that EAFormer can indeed improve the segmentation performance of the baseline model. However, as aforementioned, the annotations of COCO\_TS and MLT\_S are not accurate enough, which may make the experimental results on these two datasets unconvincing. In order to solve this problem, we have re-annotated the training, validation, and test sets of COCO\_TS and MLT\_S. Through experiments, we observe that the proposed EAFormer can still achieve better performance than previous methods with more accurate pixel-level annotations.

In summary, the contributions of this paper are as follows:
\begin{itemize}
    \item For better segmentation performance in text-edge areas, we propose Edge-Aware Transformers (EAFormer) to explicitly predict text edges and use them to guide the following encoder.
    \item Considering the low-quality annotations of COCO\_TS and MLT\_S, we have re-annotated them for experiments to make the experimental results of EAFormer on these two datasets more convincing.
    \item Extensive experiments on six scene text segmentation benchmarks demonstrate that the proposed EAFormer can achieve state-of-the-art performance and perform better in text-edge areas.
\end{itemize}

\section{Related Work}
\subsection{Scene Text Detection}

Existing scene text detection methods can be divided into two categories: regression-based methods and segmentation-based methods. Regression-based methods~\cite{ma2018arbitrary,zhou2017east,he2017deep,he2018multi} regard text detection as a distinct object detection task, where the goal is to locate text regions by predicting the offsets from anchors or pixels. However, texts exhibit significant variations in scale and orientation compared to general objects. To handle oriented texts, EAST~\cite{zhou2017east} directly regresses offsets from boundaries in an anchor-free manner. While regression-based methods perform well for quadrilateral texts, they struggle to adapt to texts with arbitrary shapes. Segmentation-based methods~\cite{liao2020real,shi2017detecting,tang2019seglink++,lyu2018multi} consider text detection as a dense binary prediction task. DBNet~\cite{liao2020real} introduces differentiable binarization within a segmentation network, allowing for adaptive threshold prediction. Although various text detection methods have been proposed, we only employ a light-weighed backbone to detect text areas in our method. Although some methods~\cite{chen2012edge,yin2023novel} in general segmentation field have proposed to introduce the edge information to improve the performance, they are not perfectly suitable for the text segmentation task, which may result from two reasons: 1) To detect edges accurately, most of them need the annotations of edges, which is time-consuming and labor-intensive. 2) Directly employing them to solve text segmentation may introduce some edges of non-text areas, leading to subpar performance. 

\subsection{Semantic Segmentation}
Semantic segmentation is a fundamental task in computer vision, which involves classifying each pixel in input images. Fully convolutional networks (FCN), which can learn dense prediction efficiently, were previously mainstream for the semantic segmentation task.  To capture contextual information at multiple scales, several methods~\cite{zhao2017pyramid,chen2018encoder} introduce dilated convolutions or spatial pyramid pooling to enlarge the receptive field. Subsequently, attention mechanisms were introduced to better capture long-range dependencies~\cite{fu2020scene,yu2020context,zheng2021rethinking}. Recently, a Transformer-based semantic segmentation method SegFormer~\cite{xie2021segformer} proposes to combine hierarchical Transformer encoders with a lightweight MLP decoder. Due to its outstanding performance, we adopt it as the baseline model of our method.

\subsection{Scene Text Segmentation}
Scene text segmentation aims at predicting fine-grained masks for texts in scene images. In the past, text segmentation methods often rely on thresholding~\cite{otsu1979threshold,su2010binarization,sauvola2000adaptive,mustafa2018binarization} or low-level features~\cite{bai2014seed,liu2006multiscale,wu2016learning,vo2018binarization} to binarize scene text images. However, these approaches often struggle with text images that have complex colors and textures, leading to poor performance. Recently, deep learning-based text segmentation methods have emerged. For instance, SMANet~\cite{bonechi2019coco_ts} adopts the encoder-decoder structure and introduces a new multi-scale attention module for scene text segmentation. TextFormer~\cite{wang2023textformer} introduces a text decoder into the hierarchical segmentation framework to enhance its ability to perceive text details. Due to the low labeling quality of previous datasets, TexRNet~\cite{xu2021rethinking} proposes the TextSeg dataset with fine-grained annotations, which contain word- and character-level bounding polygons, masks, and transcriptions. Considering the lack of Chinese texts in text segmentation, a bilingual text segmentation dataset called BTS~\cite{xu2022bts} has been proposed. The authors of BTS also developed PGTSNet, which employs a pre-trained text detection model to constrain text segmentation on detected text areas. 

\section{Methodology}
In this section, we introduce the proposed EAFormer in detail. First, we introduce the motivation of proposing EAFormer. Then, we detail each module of EAFormer, including text edge extractor, edge-guided encoder and text segmentation decoder. Finally, we introduce the loss function of our method.

\begin{figure*}[t]
  \centering
   \includegraphics[width=0.8\linewidth]{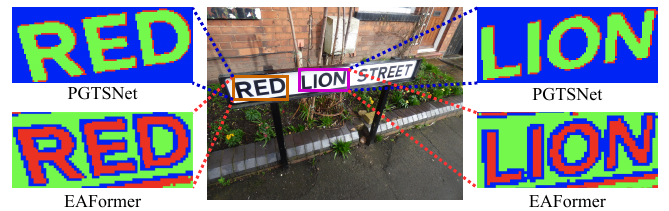}

   \caption{Feature clustering results of PGTSNet and EAFormer. The visualization indicates that PGTSNet can hardly well perceive text edges compared with EAFormer.}
   \label{fig:motivation}
\end{figure*}

\subsection{Motivation}
It is indisputable that text edges are crucial for the scene text segmentation task, especially for its downstream tasks like text erasing. Accurately segmenting text edges can provide more background information for the text erasing model to fill text areas. As shown in Figure~\ref{fig:intro}, we utilize a pre-trained inpainting model, taking different types of text masks as input, to erase texts in scene images. Through experiments, we observe that the text bounding-box mask is too coarse to provide more backgrounds for the inpainting model. In addition, the text mask with inaccurate edge segmentation makes the inpainting model mistakenly regard the pixels belonging to texts as backgrounds, resulting in poor erasing results. Only when the text mask with accurate edge segmentation is provided, the inpainting model can generate satisfactory text erasing results.

Although PGTSNet~\cite{xu2022bts} has realized the significance of text edges and employed a binary cross-entropy loss for detecting pixels at the text edges, it fails to explicitly introduce the easily available text edge information as one of the input information. In order to verify its ability to perceive text edges, we perform $K$-Means clustering on the features output by the backbone, where $K$ is set to 3, representing the background, text edge, and text center, respectively. Through the visualization results shown in Figure~\ref{fig:motivation}, we observe that this method still has certain deficiencies in perceiving text edges.

In addition, we find that the traditional edge detection algorithm can obtain accurate text edges, which may benefit the scene text segmentation task. However, since the traditional edge detection method cannot distinguish text areas and non-text areas, most edges are detected in non-text areas. If the edge detection results are directly utilized as input to assist text segmentation, it could potentially confuse the text segmentation model and adversely impact its performance. More discussions are in Section~\ref{discussions}. In the following subsection, we will introduce how our method leverages the results of traditional edge detection algorithms to achieve better performance in the text segmentation task.

\begin{figure*}[t]
  \centering
   \includegraphics[width=1.0\linewidth]{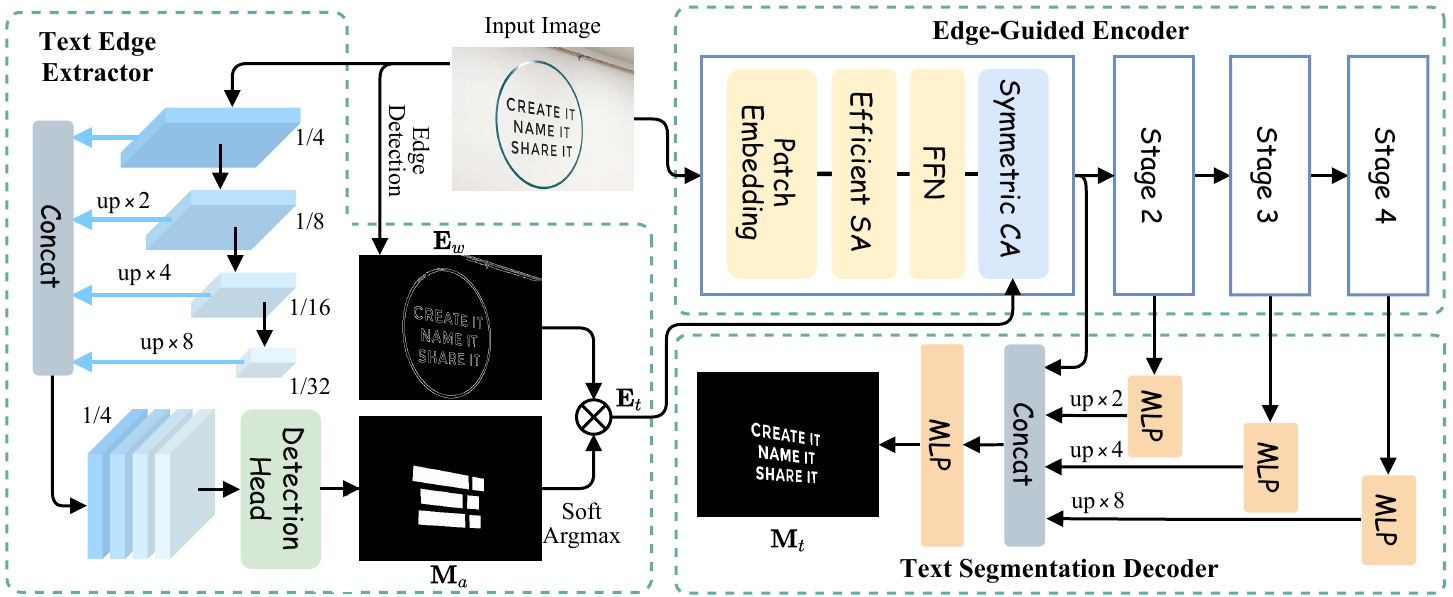}

   \caption{Overall structure of EAFormer. EAFormer consists of three modules: text edge extractor, edge-guided encoder and text segmentation decoder. `SA', `CA', and `FFN' represent self-attention, cross-attention, and feed-forward network, respectively.}
   \label{fig:overview}
\end{figure*}

\subsection{Edge-Aware Transformers (EAFormer)}
As shown in Figure~\ref{fig:overview}, the proposed EAFormer consists of three modules: text edge extractor, edge-guided encoder, and text segmentation decoder. Given the input scene text image $\mathbf{X} \in \mathbb{R}^{3\times H\times W}$, the text edge extractor is used to obtain the edges of text areas $\mathbf{E}_{t}$. Then, the text image $\mathbf{X}$ and detected text edges $\mathbf{E}_{t}$ are input into the edge-guided encoder to extract edge-aware features. Finally, the text segmentation decoder takes the features generated by the encoder as input to produce the corresponding text mask $\mathbf{M}_t$.

\noindent\textbf{Text Edge Extractor.} Since text edges are crucial for the scene text segmentation task, we propose a text edge extractor to obtain the edges of text areas. At first, we use the traditional edge detection algorithm Canny~\cite{canny1986computational} to acquire the edges of the whole input image $\mathbf{E}_w$. As aforementioned, the edges of non-text areas in $\mathbf{E}_w$ may have negative impacts on text segmentation. Therefore, we introduce a lightweight text detection model in the text edge extractor to perform edge filtering. Specifically, we first use ResNet-like~\cite{he2016deep} backbone to extract multi-level visual features $\mathcal{F}^d=\{\mathbf{F}^d_1, \mathbf{F}^d_2, \mathbf{F}^d_3, \mathbf{F}^d_4\}$, where $\mathbf{F}^d_i\in \mathbb{R}^{C_i\times H_i \times W_i}$ represents the features at the $i$-th layer of the ResNet-like backbone (more details about the backbone of text detection are introduced in the supplementary material). Then, a text detection head is employed to predict the mask of text areas $\mathbf{M}_a$, which can be formulated as:
\begin{equation}
    \mathbf{M}_a = \text{Conv}_{1\times 1}(\text{Concat}(\{\mathbf{F}^d_1, \mathbf{F}^d_2, \mathbf{F}^d_3, \mathbf{F}^d_4\}))
\end{equation}
where $\text{Conv}_{1\times 1}$($\cdot$) and Concat($\cdot$) represent the $1\times 1$ convolution layer and the concatenation operation, respectively. With the help of the mask of text areas $\mathbf{M}_a$, we can filter out the edges of non-text areas through pixel-wise multiplication between the mask of text areas $\mathbf{M}_a$ and the detected edges $\mathbf{E}_w$. Therefore, the edges of text areas $\mathbf{E}_t$ can be obtained by:
\begin{equation}
    \mathbf{E}_{t} = \mathbf{M}_a \odot \text{SoftArgmax}(\mathbf{E}_w) 
\end{equation}
It is worth mentioning that we exert the soft argmax operation on $\mathbf{E}_w$ before multiplication since joint optimizing the text detection and segmentation branch can achieve better text detection performance. Then, the filtered text edges $\mathbf{E}_t$ are subsequently input into the following edge-guided encoder to enhance its ability to distinguish pixels around text edges. 

\noindent\textbf{Edge-Guided Encoder.} Since SegFormer~\cite{xie2021segformer} exhibits outstanding abilities in semantic segmentation, we adopt it as the basic framework of the edge-guided encoder. As shown in Figure~\ref{fig:overview}, the edge-guided encoder consists of four stages, and the filtered text edges are merged at the first stage. Each encoding stage contains three submodules: overlap patch embedding, efficient self-attention, and feed-forward network. The overlap patch embedding is used to extract local features around each patch. Then, the features are input into the self-attention layer to excavate the correlations between pixels. The vanilla self-attention layer is formulated as follows:
\begin{equation}
    \text{SA}(\mathbf{Q}, \mathbf{K}, \mathbf{V})=\text{Softmax}(\frac{\mathbf{QK}^\top}{\sqrt{d_\text{head}}})\mathbf{V},
\end{equation}
where $\mathbf{Q}$, $\mathbf{K}$, and $\mathbf{V}$ are obtained by applying different embedding layers to the same features. To reduce the computational cost, we follow~\cite{xie2021segformer} to introduce the spatial reduction operation for $\mathbf{K}$ and $\mathbf{V}$. More details of spatial reduction are shown in the supplementary material. Finally, for the $i$-th stage, a feed-forward network is employed to generate the output features $\mathbf{F}^{s}_i$. Differently, we additionally introduce a symmetric cross-attention layer after the feed-forward network of the first stage to merge the extracted edge guidance $\mathbf{E}_t$. Specifically, the symmetric cross-attention layer includes two cross-attention operations between the features $\mathbf{F}^{s}_1$ from the first stage and the edge guidance $\mathbf{E}_t$. On one hand, $\mathbf{E}_t$ is regarded as Query to extract the edge-aware visual information $\mathbf{F}^{ev}$, where $\mathbf{F}^{s}_1$ is viewed as Key and Value; on the other hand, $\mathbf{F}^{s}_1$ is used as Query to further excavate the useful text edge information $\mathbf{F}^{te}$, where $\mathbf{E}_t$ is seen as Key and Value. Therefore, the final output of the first stage $\hat{\mathbf{F}}^s_1$ can be expressed as:
\begin{equation}
\begin{aligned}
    \hat{\mathbf{F}}^s_1 & = \mathbf{F}^{ev} \oplus \mathbf{F}^{te} \oplus \mathbf{F}^s_1 \\
    \mathbf{F}^{ev} & = \text{SA}(\mathbf{E}_t, \mathbf{F}^{s}_1, \mathbf{F}^{s}_1) \\
    \mathbf{F}^{te} & = \text{SA}(\mathbf{F}^{s}_1, \mathbf{E}_t, \mathbf{E}_t) \\
\end{aligned}
\end{equation}
where SA($\cdot$) represents the aforementioned self-attention operation, $\oplus$ denotes the pixel-wise addition. Subsequently, $\hat{\mathbf{F}}^s_1$ and outputs of other stages are input into the text segmentation decoder.

\noindent\textbf{Text Segmentation Decoder.} Similar to the previous method~\cite{wang2023textformer}, we employ several MLP layers to fuse features and predict the final text masks $\mathbf{M}_t$. First, we unify the channel dimension of the outputs of four stages through corresponding MLP layers. Then, they are up-sampled into the same resolution and further fused by an MLP layer. Finally, the fused features are used to predict the text masks. Assuming that the resolution of features from the $i$-th stage is $H_i \times W_i \times C_i$, the decoding process can be formulated as:
\begin{equation}
    \begin{aligned}
        \tilde{\mathbf{F}^s_i} & = \text{MLP}(C_i, C_1)(\mathbf{F}), \ \mathbf{F}\in \{\hat{\mathbf{F}}^s_1, \mathbf{F}^s_2, \mathbf{F}^s_3, \mathbf{F}^s_4\} \\
        \tilde{\mathbf{F}^s_i} & = \text{UpSample}(H_1, W_1)(\tilde{\mathbf{F}^s_i}), \ i\in \{1, 2, 3, 4\} \\
        \mathbf{F}^s & = \text{Fuse}(\mathcal{F}), \ \mathcal{F} = \{\tilde{\mathbf{F}^s_1}, \tilde{\mathbf{F}^s_2}, \tilde{\mathbf{F}^s_3}, \tilde{\mathbf{F}^s_4}\}\\
        \mathbf{M}_t & = \text{MLP}(C_1, 2)(\mathbf{F}^s)
    \end{aligned}
\end{equation}
where MLP($C_\text{in}$, $C_\text{out}$)($\cdot$) represents that the channels of input and output features in MLP are $C_\text{in}$ and $C_\text{out}$, respectively. Fuse($\cdot$) denotes that the input features are first concatenated and then reduced in the channel dimension through an MLP layer.

\subsection{Loss Function}
Previous text segmentation methods~\cite{xu2022bts,xu2021rethinking} tend to introduce various losses to improve the performance, which may bring difficulties in choosing appropriate hyper-parameters. In the proposed EAFormer, only two cross-entropy losses, text detection loss $\mathcal{L}_{det}$ and text segmentation loss $\mathcal{L}_{seg}$, are used for optimization, which can be expressed by:
\begin{equation}
    \mathcal{L} = \underbrace{\text{CE}(\mathbf{M}_t, \hat{\mathbf{M}_t})}_{\mathcal{L}_{seg}} +\ \lambda \ \underbrace{\text{CE}(\mathbf{M}_a, \hat{\mathbf{M}_a})}_{\mathcal{L}_{det}} 
\end{equation}
where $\lambda$ is the hyper-parameter to balance $\mathcal{L}_{det}$ and $\mathcal{L}_{seg}$; $\hat{\mathbf{M}}_a$ and $\hat{\mathbf{M}}_t$ are the ground truth of $\mathbf{M}_a$ and $\mathbf{M}_t$, respectively. Please note that the utilized bounding box-level supervision for $\mathbf{M}_a$ can be obtained from the semantic-level annotations, which means that the proposed method only requires the semantic-level annotations as the same as previous methods.

\section{Experiments}

\begin{table*}[t]
    \centering
    \renewcommand{\arraystretch}{1.0}
    \scalebox{1.0}{
    \begin{tabular}{ccccccc}
    \toprule
    Dataset & Images & Train/Test/Valid & Words & Classes & Languages \\
    \midrule
    ICDAR13 FST~\cite{karatzas2013icdar} & 462 & 229/233/-  & 1,944 & 36 & English \\
    COCO\_TS~\cite{bonechi2019coco_ts} & 14,690 & 11,882/2,808/-  & 139,034 & 36 & English \\
    MLT\_S~\cite{bonechi2020weak} & 6,896 & 5,540/1,356/- & 30,691 & 36 & English \\
    Total-Text~\cite{ch2017total} & 1,555 & 1,255/300/-  & 9,330 & 36 & English \\
    TextSeg~\cite{xu2021rethinking} & 4,024 & 2,646/1,038/340  & 15,691 & 36 & English \\
    BTS~\cite{xu2022bts} & 14,287 & 10,191/1,366/2,730  & 44,385 & 3,988 & Chinese,English \\
    \bottomrule
    \end{tabular}}
   \caption{Statistical details about the adopted six text segmentation datasets. }
    \label{datasets_statistics}
\end{table*}

\subsection{Datasets}
In this paper, we have conducted extensive experiments on six text segmentation benchmarks, including five English text segmentation datasets (ICDAR13 FST~\cite{karatzas2013icdar}, COCO\_TS~\cite{bonechi2019coco_ts}, MLT\_S~\cite{bonechi2020weak}, Total-Text~\cite{ch2017total}, and TextSeg~\cite{xu2021rethinking}) and one bi-lingual text segmentation dataset BTS~\cite{xu2022bts}. Some statistical details about each dataset are shown in Table~\ref{datasets_statistics}. The examples of each dataset are displayed in the supplementary material. 

\begin{figure}[t]
  \centering
   \includegraphics[width=1.0\linewidth]{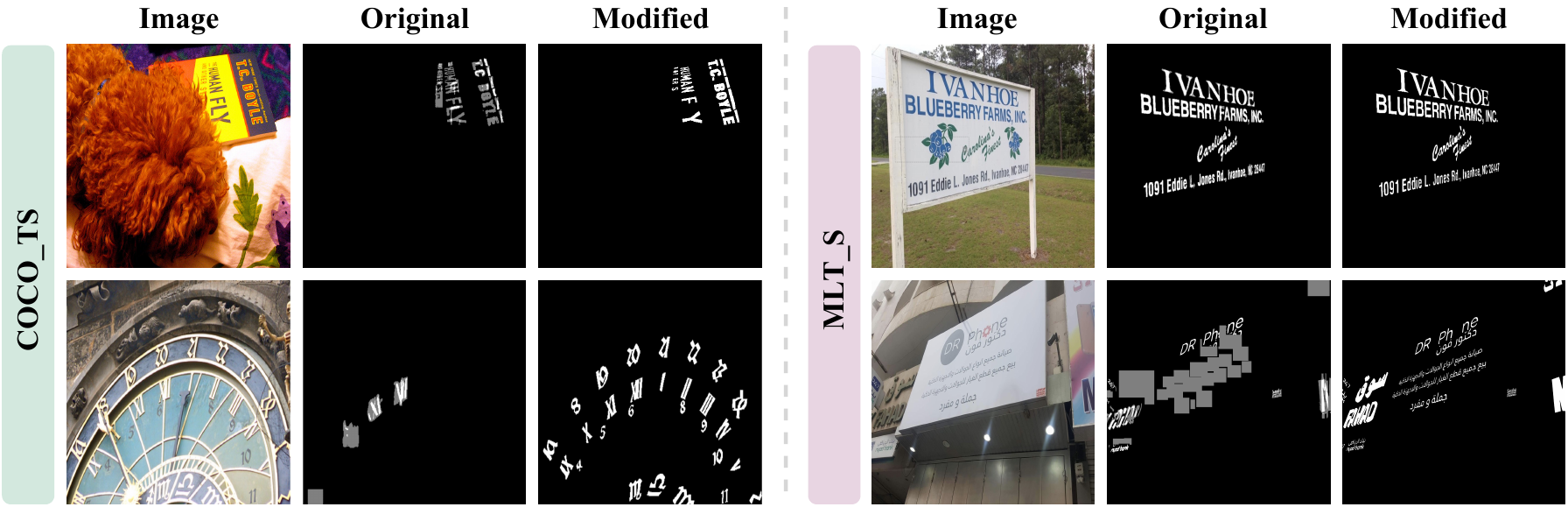}

   \caption{Comparison between original and modified annotations. The original datasets have the problems of missing and inaccurate annotations. Using re-annotated datasets to train the proposed method makes the experimental results more convincing.}
   \label{fig:anno_cmp}
\end{figure}

As shown in Figure~\ref{fig:anno_cmp}, the original annotations of COCO\_TS and MLT\_S are too coarse to train a text segmentation model with satisfactory performance. Even if the proposed method achieves better performance on these datasets, it is not sufficient to demonstrate the effectiveness of our method. To make the experimental results more convincing, we have re-annotated all samples of these two datasets and use the newly annotated datasets to conduct experiments. Figure~\ref{fig:anno_cmp} shows the comparison between original and modified annotations. 

\subsection{Implementation Details}
The proposed method is implemented with PyTorch, and we conduct all experiments on 8 NVIDIA RTX 4090 GPUs. The AdamW~\cite{loshchilov2018decoupled} optimizer is adopted with an initial learning rate 6$\times 10^{-5}$ in all experiments, and the weight decay is set to 0.01. The batch size is set to 4. Following the previous methods~\cite{wang2023textformer,ren2022looking,xu2022bts}, we also adopt some data augmentation operations, such as random cropping and flipping, in the training phase. Different from existing methods employing pre-trained models to detect text areas~\cite{xu2022bts} or recognize characters~\cite{xu2021rethinking}, all modules in the proposed EAFormer are jointly trained. In other words, no additional datasets are used when training EAFormer. Two thresholds of Canny are set to 100 and 200, respectively. To evaluate the performance of the proposed method, we utilize both foreground Intersection-over-Union (fgIoU) and F-score on foreground pixels. The metrics of fgIoU and F-score follow the percentage format and decimal format, respectively.

\begin{table*}[t]
    \centering
    \renewcommand{\arraystretch}{1.0}
    \scalebox{0.89}{
    \begin{tabular}{l|cc|cc|cc|cc|cc}
    \toprule
    \multirow{2}*{Method} & \multicolumn{2}{c|}{ICDAR13 FST} & \multicolumn{2}{c|}{COCO\_TS} &  \multicolumn{2}{c|}{MLT\_S} & \multicolumn{2}{c|}{Total-Text} & \multicolumn{2}{c}{TextSeg} \\
    ~ & fgIoU & F-score & fgIoU & F-score & fgIoU & F-score & fgIoU & F-score & fgIoU & F-score \\
    \midrule
    PSPNet~\cite{zhao2017pyramid,bonechi2019coco_ts}  & - & 0.797 & - & - & - & - & - & 0.740 & - & - \\
    SMANet~\cite{bonechi2020weak}  & - & 0.785 & - & - & - & - & - & 0.770 & - & - \\
    DeeplabV3+~\cite{chen2018encoder}  & 69.27 & 0.802 & 72.07 & 0.641 & 84.63 & 0.837 & 74.44 & 0.824 & 84.07 & 0.914 \\
    HRNetV2-W48~\cite{wang2020deep}  & 70.98 & 0.822 & 68.93 & 0.629 & 83.26 & 0.836 & 75.29 & 0.825 & 85.03 & 0.914 \\
    HRNet-OCR~\cite{yuan2020object}  & 72.45 & 0.830 & 69.54 & 0.627 & 83.49 & 0.838 & 76.23 & 0.832 & 85.98 & 0.918 \\
    TexRNet~\cite{xu2021rethinking} & \textbf{73.38} & 0.850 & 72.39 & 0.720 & 86.09 & 0.865 & 78.47 & 0.848 & 86.84 & 0.924 \\
    ARM-Net~\cite{ren2022looking} & - & \textbf{0.851} & - & - & - & - & - & 0.854 & - & 0.927\\
    TFT~\cite{yu2023scene} & \underline{72.71} & \underline{0.845} & \underline{73.40} & \underline{0.847} & \underline{87.80} & \underline{0.935} & \underline{82.10} & \underline{0.902} & 87.11 & 0.931 \\
    TextFormer~\cite{wang2023textformer} & 72.27 & 0.838 & 73.20 & 0.745 & 86.66 & 0.908 & 81.56 & 0.887 & \underline{87.42} & \underline{0.933} \\
    \midrule
    SegFormer (Base)~\cite{xie2021segformer} & 60.44 & 0.753 & 63.17 & 0.774 & 78.77 & 0.863 & 73.31 & 0.846 & 84.59 & 0.916 \\
    EAFormer (Ours) & 72.63 & 0.840 & \textbf{81.03} & \textbf{0.895} & \textbf{89.02} & \textbf{0.942} & \textbf{82.73} & \textbf{0.906} & \textbf{88.06} & \textbf{0.939} \\
    \bottomrule
    \end{tabular}}
    \caption{Performance comparison on five English text segmentation datasets. The bold and underlined numbers represent the best and second-best results, respectively.}
    \label{english-table}
\end{table*}

\begin{table}[t]
    \centering
    \renewcommand{\arraystretch}{1.0}
    \scalebox{1.0}{
    \begin{tabular}{l|cc|cc}
    \toprule
    \multirow{2}*{Method} & \multicolumn{2}{c|}{COCO\_TS} & \multicolumn{2}{c}{MLT\_S}\\
    ~ & fgIoU & F-score & fgIoU & F-score\\
    \midrule
    TextFormer~\cite{wang2023textformer} & 52.73 & 0.688 & 74.83 & 0.861 \\
    EAFormer (Ours) & \textbf{64.82} & \textbf{0.786} & \textbf{81.92} & \textbf{0.900} \\
    \bottomrule
    \end{tabular}}
    \caption{The experimental results on re-annotated COCO\_TS and MLT\_S. The results of TextFormer are obtained by re-implementing the method. 
}
    \label{table-reanno}
\end{table}

\subsection{Experimental Results}
\textbf{Quantitative Comparison.} To comprehensively evaluate EAFormer, we have conducted experiments on both English and bi-lingual text segmentation datasets. The experimental results on five English text segmentation datasets are shown in Table~\ref{english-table}. Compared with previous methods, EAFormer can achieve a clear improvement in fgIoU and F-score on most benchmarks. For example, on TextSeg, EAFormer outperforms the previous SOTA method TextFormer~\cite{wang2023textformer} by 0.64\% and 0.6\% in fgIoU and F-score, respectively. Although the original COCO\_TS and MLT\_S datasets have coarse annotations, the proposed EAFormer can still exhibit better performance, such as achieving a 7.63\% improvement in fgIoU on the COCO\_TS dataset compared with TFT~\cite{yu2023scene}. Considering that the experimental results based on inaccurate annotations are not convincing enough, we have re-annotated both the training dataset and the test dataset of COCO\_TS and MLT\_S. The experimental results based on re-annotated datasets are displayed in Table~\ref{table-reanno}. Through experiments, we observe that the proposed EAFormer can still achieve a considerable performance improvement when the datasets with more accurate annotations are used for training and test. Compared with the results on original datasets, the performance on re-annotated datasets seems to drop a lot. The following two reasons may explain this phenomenon: 1) There are many blurred texts in the datasets, which indeed bring certain challenges to the model to deal with text edges; 2) The re-annotated test datasets are more accurate and there are no ignored areas in evaluation.   In addition, we also conduct experiments on the bi-lingual text segmentation dataset BTS~\cite{xu2022bts}, and the results are shown in Table~\ref{bts-table}. Although PGTSNet unfairly introduces a pre-trained text detector, EAFormer can still achieve an improvement of 1.6\%/2.8\% in fgIoU/F-score, which validates the effectiveness of the proposed method.

Since we introduce a light-weighted text detection head, it is inevitable to introduce more parameters. We have evaluated the number of parameters and inference speed. Compared with the previous SOTA method TextFormer (85M parameters and 0.42s per image), the proposed model has 92M parameters and costs 0.47s per image in average. With slight increasing in the number of parameters, our method can achieve significant performance improvements.

\begin{table}[t]
    \centering
    \renewcommand{\arraystretch}{1.0}
    \scalebox{1.0}{
    \begin{tabular}{l|cc}
    \toprule
    \multirow{2}*{Method} & \multicolumn{2}{c}{BTS}\\
    ~ & fgIoU & F-score \\
    \midrule
    DeeplabV3+~\cite{chen2018encoder} & 71.15 & 0.796 \\
    HRNetV2-W48~\cite{wang2020deep} & 81.84 & 0.861  \\
    HRNetV2-W48+OCR~\cite{yuan2020object} & 82.76 & 0.866 \\
    TexRNet (DeeplabV3+, w/o Cls)~\cite{xu2021rethinking} & 83.68 & 0.883 \\
    TexRNet (DeeplabV3+, w/ Cls)~\cite{xu2021rethinking} & 83.81 & 0.892  \\
    PGTSNet~\cite{xu2022bts} & 86.48 & 0.909 \\
    TFT~\cite{yu2023scene} & \underline{87.84} & \underline{0.935} \\
    TextFormer~\cite{wang2023textformer} & 86.97 & 0.930 \\
    \midrule
    SegFormer (Baseline)~\cite{xie2021segformer} & 84.99 & 0.908 \\
    EAFormer (Ours) & \textbf{88.08} & \textbf{0.937} \\
    \bottomrule
    \end{tabular}}
    \caption{Performance comparison on BTS. The bold and underlined numbers represent the best and second-best results, respectively.}
    \label{bts-table}
\end{table}

\begin{figure*}[ht]
  \centering
  \includegraphics[width=0.99\linewidth]{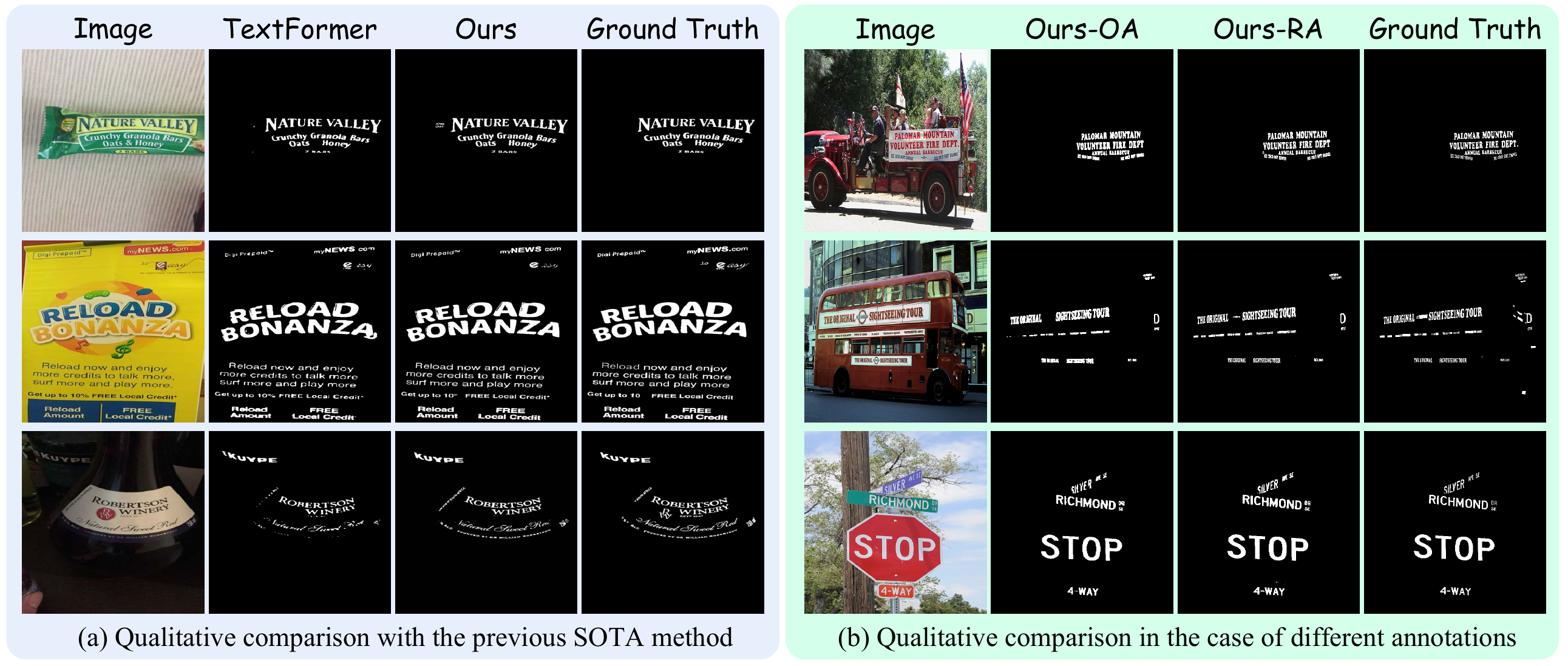}

   \caption{Visualizations of qualitative comparison between different methods or training with different annotations. `OA' and `RA' indicate training EAFormer with original annotations and re-annotations, respectively.}
   \label{fig:res}
\end{figure*}

\noindent \textbf{Qualitative Comparison.} We also compare EAFormer with previous methods in term of segmentation quality through visualizations. As shown in Figure~\ref{fig:res}, the proposed EAFormer can perform better than previous methods at text edges, which benefits from the introduced edge information. In addition, for COCO\_TS and MLT\_S, we compare the segmentation results based on both the original and modified annotations. Although Table~\ref{table-reanno} indicates that the performance of our method has declined when using the re-annotated datasets for training and test, the visualizations in Figure~\ref{fig:res} demonstrate that our model is able to achieve better segmentation results based on the re-annotated datasets. More visualizations are shown in the supplementary material.

\begin{table}[t]
    \centering
    \renewcommand{\arraystretch}{1.1}
    \scalebox{1.0}{
    \begin{tabular}{c|cc|cc}
    \toprule
    \multirow{2}*{$\lambda$} & \multicolumn{2}{c|}{TextSeg} & \multicolumn{2}{c}{BTS}\\
    ~ & fgIoU & F-score & fgIoU & F-score\\
    \midrule
    0.1 & 84.03 & 0.910 & 86.45 & 0.913 \\
    0.5 & 87.33 & 0.926 & 87.03 & 0.931 \\
    1.0 & \textbf{88.06} & \textbf{0.939} & \textbf{88.08} & \textbf{0.937} \\
    5.0 & 87.67 & 0.934 & 87.62 & 0.935 \\
    10.0 & 87.94 & 0.937 & 87.58 & 0.933 \\
    \bottomrule
    \end{tabular}}
    \caption{The experimental results of choosing $\lambda$. When $\lambda$ is set to 1.0, the proposed method can achieve the best performance.}
    \label{lambda-choice}
\end{table}

\subsection{Ablation Study}
\textbf{Hyper-parameter $\lambda$.} During training EAFormer, two losses are used for optimization. The hyper-parameter $\lambda$ is to balance the weights of $\mathcal{L}_{det}$ and $\mathcal{L}_{seg}$, and an appropriate $\lambda$ may result in better performance. Thus, we conduct some experiments to select $\lambda$ ranging from $\{0.1, 0.5, 1.0, 5.0, 10.0\}$, and the experimental results are shown in Table~\ref{lambda-choice}. When $\lambda$ is set to 1.0, EAFormer reaches the best and it boosts fgIoU/F-score by 3.47\%/2.3\% compared with the baseline model on TextSeg. The results shown in Table~\ref{lambda-choice} indicate that $\lambda$ has little impact on performance when it ranges from \{0.5, 1.0, 5.0, 10.0\}. However, if $\lambda$ is set to 0.1, the performance of EAFormer is unsatisfactory, which may result from that too small $\lambda$ makes it difficult for the text detection module to converge and further affects the performance of text segmentation. Therefore, we set $\lambda$ to 1.0 for all experiments in this paper.

\begin{table}[t]
    \centering
    \renewcommand{\arraystretch}{1.1}
    \scalebox{1.0}{
    \begin{tabular}{cc|cc|cc}
    \toprule
    
    \multirow{2}*{EF} & \multirow{2}*{EG} & \multicolumn{2}{c|}{TextSeg} & \multicolumn{2}{c}{BTS} \\
    
    ~ & ~ & fgIoU & F-score & fgIoU & F-score \\
    \midrule
    & & 84.59 & 0.916 & 84.99 & 0.908 \\
    \ding{51} &  & 86.85 & 0.927 & 87.35 & 0.922\\
     & \ding{51} & 81.03 & 0.832 & 80.35 & 0.828\\
    \ding{51} & \ding{51} & 88.06 & 0.939 & 88.08 & 0.937\\
    \bottomrule
    \end{tabular}}
    \caption{The experimental results of ablation studies. If only the edge guidance is introduced, the edges of non-text areas have negative impacts on the proposed method.}
    \label{ablation}
\end{table}

\noindent \textbf{Edge Filtering and Edge Guidance.} In the proposed EAFormer, the edge filtering in the text edge extractor and the edge guidance in the edge-guided encoder are two key components. To evaluate the performance gain of these two strategies, we conduct ablation experiments on them and the results are shown in Table~\ref{ablation}. Please note that when only edge filtering is used, the extracted edge information is concatenated with the input image and fed into the SegFormer-based encoder. As shown Table~\ref{ablation}, introducing edge filtering can lead to a clear performance improvement. However, if only edge guidance is introduced, the performance of our method is subpar. A possible reason is that the edges of non-text areas introduce more interference information, resulting in that the model cannot effectively use the extracted edges to assist text segmentation. Therefore, both edge filtering and edge guidance are necessary for our method, and when both of them are adopted, EAFormer can achieve the SOTA performance.

\section{Discussions}
\label{discussions}

\textbf{Filtering out the edges of non-text areas.} In the text edge extractor module, we propose to filter out edge information in non-text areas to avoid their adverse impact on model performance. In the section of the ablation experiment, we can know that filtering the edge information of non-text areas can clearly improve the performance. Through visualizations (see the supplementary material), we observe that when all edge information is used to assist segmentation, the model will mistakenly believe that areas with edge information should be classified as foreground. Therefore, in order to give the model explicit edge guidance, the proposed method only retains the edge information of the text area as input.

\begin{figure*}[t]
  \centering
   \includegraphics[width=0.99\linewidth]{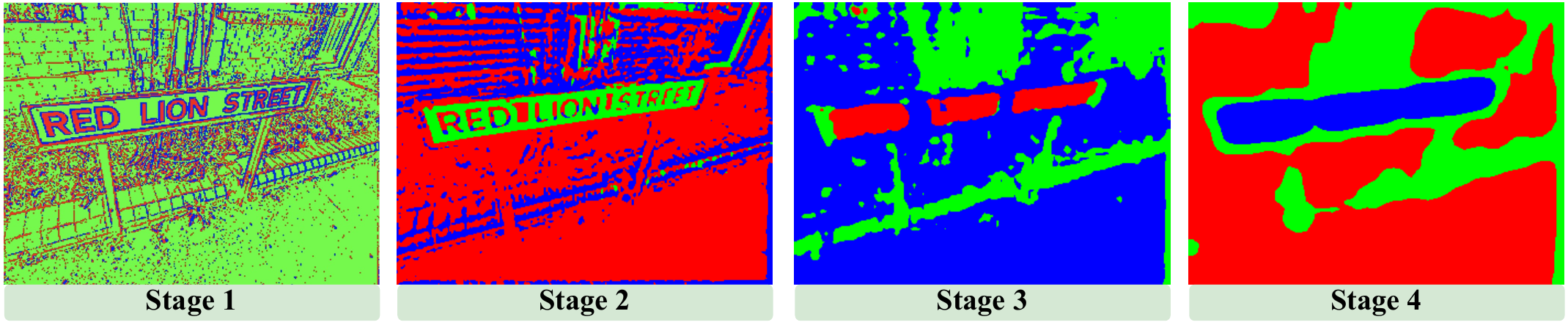}

   \caption{The clustering results of features from different stages. It is reasonable to introduce the text-edge guidance at the first stage since only the features of the first stage focus on edge information.}
   \label{fig:fist-stage}
\end{figure*}

\noindent \textbf{Introducing text edges at different layers.} In the edge-guided encoder, we extract edge-enhanced feature information only in the first stage through symmetric cross-attention. It is well known that lower-level features are more sensitive to text edge information. We have visualized the clustering results of features at different stages in Figure~\ref{fig:fist-stage}, and the visualizations indicate that only the features of the first stage focus on the edge information. Therefore, it is reasonable and effective to introduce detected edges at an earlier stage. We also tried to introduce the edge guidance at other stages to conduct experiments (detailed results are shown in the supplementary material). The experimental results indicate that the higher the stage at which the detected edges are introduced, the smaller the performance improvement of EAFormer. In particular, when the detected edges are introduced at the third or fourth stage, the performance of EAFormer is even lower than the baseline.

\noindent \textbf{Utilizing an off-the-shelf text detector.} In the text edge extractor, we employ a lightweight text detector consisting of a ResNet-based backbone and an MLP decoder. In fact, we can utilize an off-the-shelf text detector that has been pre-trained on text detection datasets, which can help EAFormer achieve better performance in practical applications. Since this may be unfair to the previous method, we only explore the performance upper limit of EAFormer. In experiments, using pre-trained DBNet~\cite{liao2020real} to replace the lightweight text detector module, the performance of EAFormer on TextSeg can reach a new SOTA performance (90.16\%/95.2\% in fgIoU/F-score).

\noindent \textbf{Differences from previous edge-guided methods.} Actually, the incorporation of edge information for segmentation is a well-explored strategy~\cite{ma2021boundary,cong2022boundary,liu2022edge}. However, our method still has some differences from previous works. First, BCANet~\cite{ma2021boundary} and BSNet~\cite{cong2022boundary} need edge supervision while the proposed method directly employs Canny to extract edges. Although EGCAN~\cite{liu2022edge} also uses Canny, our method additionally introduce edge filtering to reserve useful edge information, which is specifically designed for text segmentation. In addition, EGCAN fuses the edge information in all encoder layers while our method fuses the edge information only in the first layer through our designed symmetric cross-attention.

\noindent \textbf{Limitations.} To filter the edges of non-text areas, we introduce a light-weighted text detector, which may slightly increase the number of learnable parameters. In addition, we only utilize the off-the-shelf edge detection algorithm Canny to extract text edges rather than employing a better deep-learning-based edge detection method. Introducing a SOTA edge detection method may further improve the performance of our method.

\section{Conclusion}
In this paper, we propose Edge-Aware Transformers, called EAFormer, to solve inaccurate text segmentation at text edges. Specifically, the traditional edge detection algorithm Canny is employed to extract edges. To avoid involving the edges of non-text areas, a light-weighted text detection module is adopted to filter out the useless edges for segmenting texts. In addition, based on SegFormer, we propose an edge-guided encoder to enhance its ability to perceive text edges. Considering that the low-quality annotations of several datasets may affect the credibility of experimental results, we have re-annotated these datasets. Extensive experiments are conducted on publicly available benchmarks, and the SOTA results validate the effectiveness of EAFormer in the text segmentation task.

\section*{Acknowledgements}
This work was supported in part by the National Natural Science Foundation of China (No.62176060), STCSM project (No.22511105000), Shanghai Municipal Science and Technology Major Project (No.2021SHZDZX0103), and the Program for Professor of Special Appointment (Eastern Scholar) at Shanghai Institutions of Higher Learning.

% ---- Bibliography ----
%
% BibTeX users should specify bibliography style 'splncs04'.
% References will then be sorted and formatted in the correct style.
%
\bibliographystyle{splncs04}
\bibliography{main}
\end{document}